%% file: main.tex
\documentclass[letterpaper, 10 pt, conference]{ieeeconf}  

\IEEEoverridecommandlockouts                              
\usepackage{bmathshortcuts}

\usepackage{algorithmic}
\usepackage[ruled,vlined]{algorithm2e}
\usepackage{amsmath,amssymb,amsfonts}
\usepackage{bm}
\usepackage{booktabs}
\usepackage{cite}
\usepackage{color}
\usepackage{comment}
\usepackage{fancyhdr}
\usepackage{float}
\usepackage{balance}
\usepackage{gensymb}
\usepackage{graphicx}
\usepackage{caption} 
\usepackage{subcaption} 
\usepackage[hidelinks]{hyperref} 
\usepackage{mathrsfs}
\usepackage{soul}
\usepackage{textcomp}
\usepackage{times}
\usepackage[usenames,dvipsnames]{xcolor} 
\usepackage{url}
\include{pythonlisting}

{\vspace{-\topsep}\begin{itemize}\itemsep1pt \parskip0pt \parsep1pt}
{\end{itemize}\vspace{-\topsep}}

{\vspace{-\topsep}\begin{enumerate}\itemsep1pt \parskip0pt \parsep1pt}
{\end{enumerate}\vspace{-\topsep}}

\begin{document}

\input{P00_Title}
\input{P0_Abstract}

\input{P1_Introduction}
\input{P2_Methodology}

\input{P3_Results}
\input{P4_Discussion}

\input{P5_Conclusion}


\bibliographystyle{ieeetr}
\bibliography{references}
\balance

\end{document}

%% file: P00_Title.tex
\title{\LARGE \bf
Maximizing Consistent Force Output for Shape Memory Alloy Artificial Muscles in Soft Robots}


\author{Meredith L. Anderson$^{1}$, Ran Jing$^{1}$, Juan C. Pacheco Garcia$^{1}$, Ilyoung Yang$^{1}$,\\Sarah Alizadeh-Shabdiz$^{1}$, Charles DeLorey$^{1}$, Andrew P. Sabelhaus$^{1}$
\thanks{This work was supported in part by the U.S. National Science Foundation CSSI program under Award No. 2209783.}
\thanks{$^1$All authors are with the Department of Mechanical Engineering, Boston University, Boston MA, USA. {\tt\small \{merland, rjing, jcp29, alviny21, sar3, deloreyc, asabelha\}@bu.edu} }
}


\maketitle
\pagestyle{empty}  
\thispagestyle{empty} 

%% file: P0_Abstract.tex
\begin{abstract}
Soft robots have immense potential given their inherent safety and adaptability, but challenges in soft actuator forces and design constraints have limited scaling up soft robots to larger sizes.
Electrothermal shape memory alloy (SMA) artificial muscles have the potential to create these large forces and high displacements, but consistently using these muscles under a well-defined model, in-situ in a soft robot, remains an open challenge.
This article provides a system for maintaining the highest-possible consistent SMA forces, over long lifetimes, by combining a fatigue testing protocol with a supervisory control system for the muscles' internal temperature state.
We propose a design of a soft limb with swap-able SMA muscles, and deploy the limb in a blocked-force test to quantify the relationship between the measured maximum force at different temperatures over different lifetimes. 
Then, by applying an invariance-based control system to maintain temperatures under our long-life limit, we demonstrate consistent high forces in a practical task over hundreds of cycles.
The method we developed allows for practical implementation of SMAs in soft robots through characterizing and controlling their behavior in-situ, and provides a method to impose limits that maximize their consistent, repeatable behavior.

\end{abstract}

%% file: P1_Introduction.tex
\section{Introduction}

Many of the most commonly-claimed benefits of soft robots revolve around their potential for human interaction \cite{laschi_soft_2016,majidi_soft_2014,sanan_physical_2011}.
However, as of yet, soft robots have generally been limited to size scales and forces much smaller than humans -- often by the limitations of soft materials and actuators.
Larger soft robots either do not transmit meaningful environmental forces\cite{takeichi_development_2017,takeichi_development_2017-1} or have design limitations for mass, speed, and control effort \cite{best2015control,Usevitch_isoperimetric_2020,li_scaling_2021}.
Similarly, higher-force soft robot limbs typically require another mechanism to generate pressure\cite{ang_design_2020,liu_soft_2021}.
In contrast, soft actuators with higher work density and mass efficiency could overcome many of these challenges, particularly shape memory materials like Nitinol wires\cite{Rich_untethered_2018}.
Yet these actuators suffer from other challenges, particularly the difficulty of modeling that prompts precision testing only\cite{zakerzadeh_modeling_2011,lee_precise_2013,ge_preisach-model-based_2021}, difficult-to-sense states\cite{bhargaw_performance_2021,sabelhaus_-situ_2022}, and inconsistent stimulus-response due to functional fatigue \cite{eggeler_structural_2004}.

\begin{figure}[t]
    \centering
    \includegraphics[width=1\columnwidth]{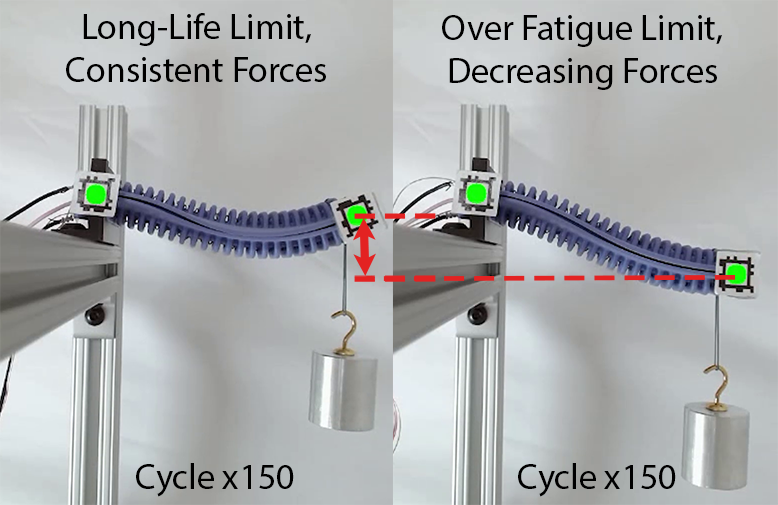}
    \caption{Example comparison of force output, in a practical task lifting a 50g mass, after 150 cycles. The SMA whose states (temperature) are controlled at proposed long-life limit generates more force and displacement than another actuated past the fatigue limit.}
    \label{fig:overview}
    \vspace{-0.3cm}
\end{figure}

This article proposes a method to generate consistent, high forces from shape memory alloy (SMA) artificial muscle actuators, in-situ in a soft robot limb.
Our approach combines a design that can ``hot swap'' the actuators and their sensors within the limb, a testing procedure to determine the actuators' cycle life, and a feedback controller to maintain the actuators' critical states below a fatigue limit.
This framework takes a significant step toward deployment of SMA actuators in their promising soft robotics applications with high force \cite{she2015design}, fast speeds \cite{Cianchetti2018biomedical}, and large deflections \cite{wertz_trajectory_2022,patterson_untethered_2020}, now adding extremely long lifetimes with predictable behavior at the highest possible forces (Fig. \ref{fig:overview}).



Degradation of actuators is a persistent problem in soft robotics, which has mostly been addressed with bespoke mechanism designs \cite{villanuevaBiomimeticRoboticJellyfish2011,coloradoBiomechanicsSmartWings2012,shin_new_2014}, empirically limiting the robot's outputs \cite{stilli_novel_2017}, or hard stops in feedback \cite{balasubramanian_fault_2020}.
SMA actuation in particular has relied upon fixed input limits \cite{kuribayashi_improvement_1991,liuReinforcementLearningControl2019,yeeharntehArchitectureFastAccurate2008,jinStarfishRobotBased2016}, which do not directly capture fatigue or degradation: since SMAs are electrothermally-actuated via electric current and Joule heating, their degradation arises from a complex interaction of thermal and mechanical cycling \cite{sofla_cyclic_2008,mohdjaniReviewShapeMemory2014,chinMachineLearningSoft2020}.

Recent work has instead developed a simple sensing architecture for temperature states in SMAs, in-situ \cite{sabelhaus_-situ_2022}, which can be used to constrain the actuators' temperature under predictive feedback control \cite{sabelhaus_safe_2022,jing2022safe}.
However, it was unknown if this approach did prevent fatigue, nor did it provide guidance on practical application.
This article proposes a comprehensive answer via a generalized procedure.

Previous work quantifying degradation and fatigue in SMA actuators has been difficult to translate to soft robotics applications.
Approaches that use constitutive models of SMAs can predict fatigue \cite{ryu2011modified, xu2021finite, mohammadgholipour2023mechanical}, but only in specialized test setups with precision instruments, up to and including microstructure measurements \cite{cisse2016review} that are impractical in most soft robots.
Similarly, grey-box models require parameter tuning \cite{copaci2020flexible} that can break once the actuator is assembled into a soft robot.
Numerical simulations are computationally intensive \cite{wang20173d}, and do not provide concise solutions that map to measurable states in hardware.
Even then, most of these techniques approximate the very prescient loading conditions in soft robots, either thermal actuation \cite{he2023comprehensive} or electrothermal coupling in multiple loading-unloading cycles \cite{woodworth2022shape}.
This article addresses the gap by proposing a practical, hardware-driven framework that has been empirically validated in a soft robot in realistic conditions.

\subsection{Contribution, Novelty, and Impact}
This article contributes a framework comprising: 

\begin{itemize}
\item An in-situ analysis protocol to determine SMA lifetime versus thermal-mechanical loading in soft robotic limbs,
\item Deployment of a supervisory feedback controller to enforce the most favorable (highest force) long-life operation of an SMA actuator,
\item Validation in a representative robot in hardware, with force and displacement preserved under feedback for hundreds of cycles in a practical task.
\end{itemize}

\noindent This methodology is, to the author's knowledge, the first deployment of a fatigue prevention system in practice for thermal or shape memory mechanisms in soft robots, and the first feedback controller that maximizes output forces while preventing that fatigue.
This system could overcome persistent challenges in time-varying modeling and short lifetimes, accelerating these robots' adoption and impact \cite{buckner_shape_2020,mazzolai_roadmap_2022} from locomotion \cite{baines_multi-environment_2022} to wearables \cite{chenal_variable_2014}.


%% file: P2_Methodology.tex
\section{Methodology}

Our approach combines a hardware design and two layers of feedback control into an experimental procedure, both to determine our long-life limit, then to maintain that limit in subsequent experiments.

\subsection{Hardware}
The soft limbs used in the test setup are adapted from prior designs \cite{sabelhaus_-situ_2022,jing2022safe,wertz_trajectory_2022}, now with a quick replacement (``hot swap'') feature for the SMA actuators and associated sensors, allowing us to fatigue then replace the actuators between tests. 
The manufacturing process casts the limb's main body using a 3D-printed mold and silicone polymer (Smooth-Sil 945, Smooth-On).
In prior designs \cite{Pacheco2023comparison}, coiled SMA wires (Dynalloy Flexinol $90^\circ$C, 0.020'' diam.) were inserted into cavities in the limb and affixed with Silpoxy adhesive, a common technique \cite{patterson_untethered_2020,patterson_robust_2022}.

\begin{figure}[t]
\vspace{0.2cm}
    \centering
    \includegraphics[width=1\columnwidth]{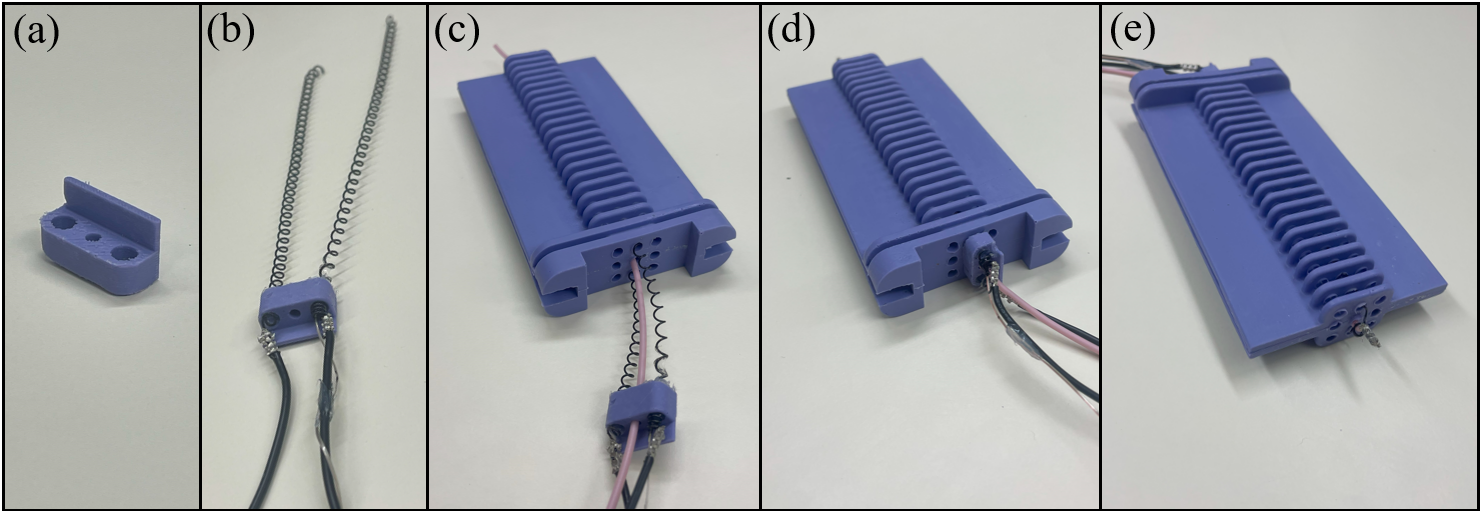}
    \caption{Limb fabrication process with the incorporation of the module (a). The SMA is crimped to a wire which is bonded to the module along with the thermocouples using Sil-Poxy. The created bundle (b) is threaded through the limb with the ground/pink wire (c). Once the bundle is inserted all the way through the limb (d), the SMAs and ground wire are crimped at the tip to secure the bundle in the limb (e).}
    \label{fig:bundle_fab}
    
\end{figure}

\begin{figure}[t]
    \centering
    \includegraphics[width=1\columnwidth]{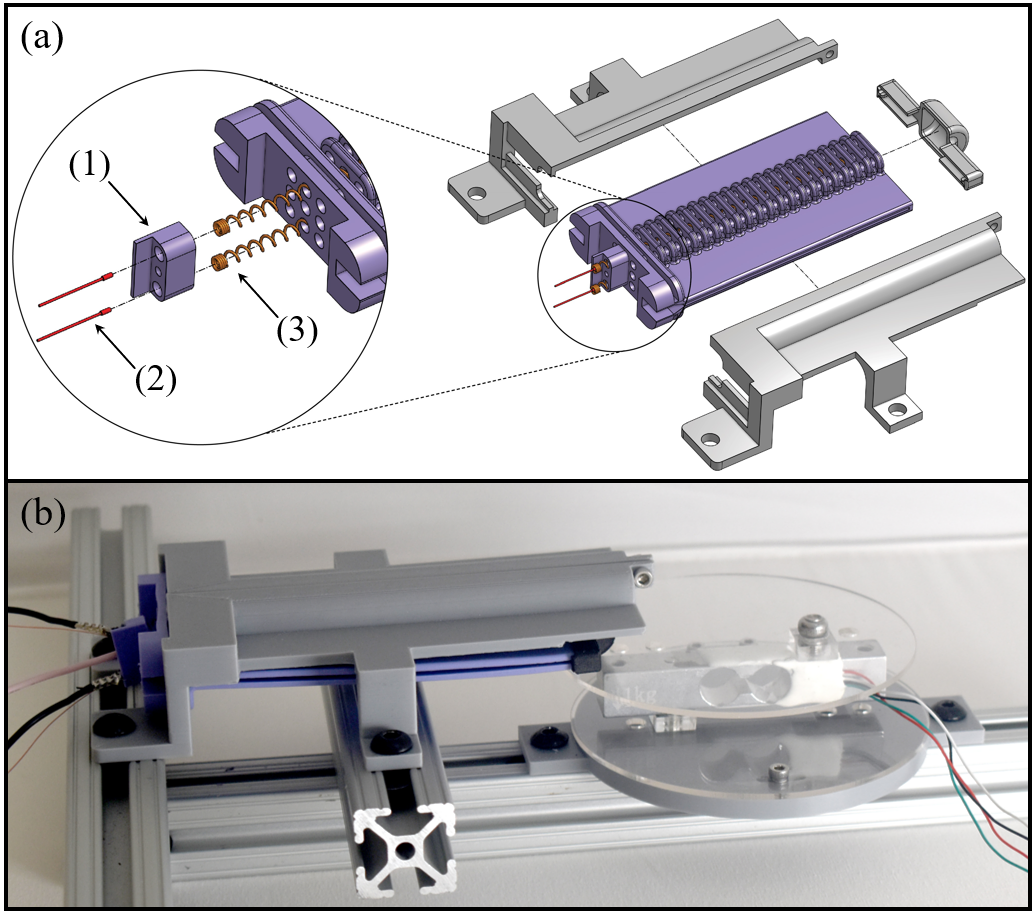}
    \caption{Testing and limb setup for recording force output. The bundle, composed of the module (1), thermocouples (2), and SMA wires (3), is placed in the center configuration of the limb then placed into a 3D printed bracket (a) that limits the movement of the limb towards the force plate. The limb is then fitted with a foot at its end (a) to localize force output and placed into contact with the force plate (b).}
    \label{fig:test_setup}
    \vspace{-0.5cm}
\end{figure}

Our proposed approach instead uses a small, replaceable, modular bracket for the SMA coils (Fig. \ref{fig:bundle_fab}(a), Fig. \ref{fig:test_setup}(a)(1)).
We attach the SMA coils to this bracket to form a ``bundle.''
For temperature measurement, a thermocouple (5TC-TT-K-36-72, Omega) is attached to each SMA on the rear of the bundle by thermally conductive, electrically insulating epoxy (Fig. \ref{fig:bundle_fab}(b), Fig. \ref{fig:test_setup}(a)(2-3)).
Crucially, we can then easily insert or remove both the actuators and sensors without compromising the limb body (Fig. \ref{fig:bundle_fab}(c)-(d)), only needing to crimp or cut an electrical connection at the limb's distal end for a return path for the current (Fig. \ref{fig:bundle_fab}(e)). 
The hot-swap procedure can be readily completed in under five minutes, and was a key enabler of our enormous dataset with $\sim$20 prototype bundles used to validate our method.

\begin{figure}[t]
    \centering
    \includegraphics[width=1\columnwidth]{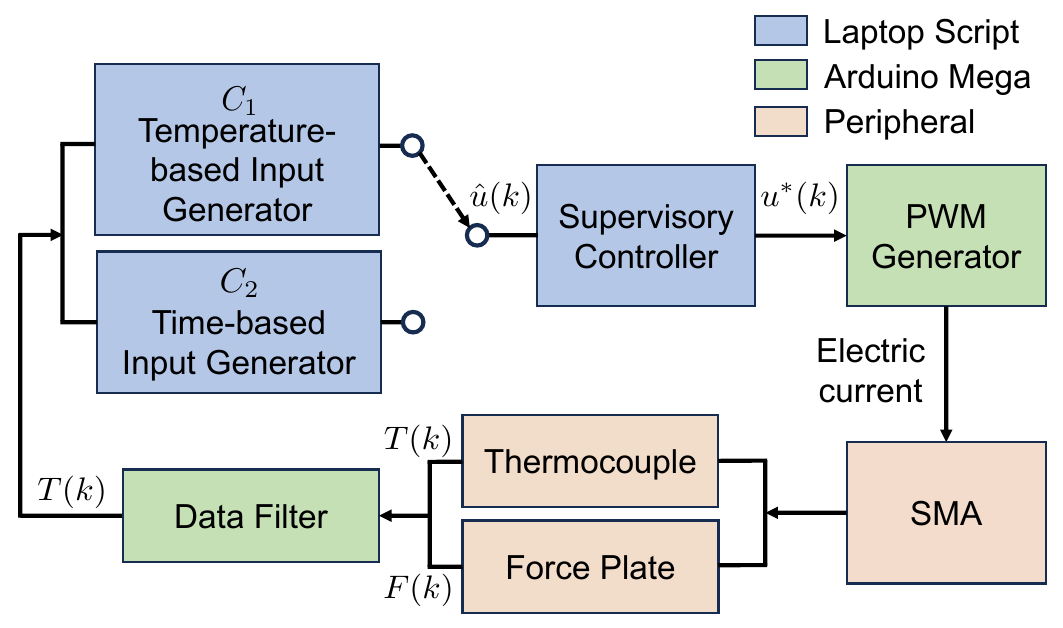}
    \caption{Logical architecture for the proposed system. The SMA temperature and force output are collected and utilized in the input generator and the supervisory controller. In each experiment trial, one of the two input generators is selected to determine when to start/stop heating in every heating cycle. The generators' PWM input is then limited by the temperature supervisor to make sure the SMA stably reaches the desired temperature. }
    \label{fig:system_ach}
    \vspace{-0.3cm}
\end{figure}

The electronics for our limb include actuation and sensing, and our test setup also integrates a force plate, as well as a computer vision system for experiments.
The SMA muscles are powered independently by Joule heating, via a 7V power supply, with one MOSFET each controlled by a pulse width modulation (PWM) voltage signal. 
A microcontroller (Arduino Mega 2560) applies this PWM as a control input to vary the effective electrical current flowing through the wire, generating heat. 
The microcontroller communicates over USB serial with a Python script running on a laptop (Core i5, 2.6GHz, 16 GB RAM), which performs the control calculations.
Temperature measurements are obtained from an amplifier (MCP9600) connected over I2C to the same microcontroller (Fig. \ref{fig:system_ach}).

To measure the force generated by the SMA, we propose a block-force test setup (Fig. \ref{fig:test_setup}) with a load cell force sensor. 
The limb is fabricated and then inserted into a 3D printed bracket above the force plate, preventing deflection upwards and making our tests more consistent.
This off-the-shelf sensor uses strain gauges to create a Wheatstone bridge, and an analog-to-digital converter (AVIA HX711) connected to the same microcontroller to record and transmit forces.

For our experiments with free space motion of the limb, the laptop Python script uses the AprilTag 2 library \cite{wang_apriltag_2016} with fiducial markers placed on the limb.


\subsection{Supervisory Controller for Temperature}

The approach in this article uses temperature feedback to maintain safe long-life operation.
To do so, we deploy our previous work's supervisory control system that dynamically saturates control inputs to ensure a temperature limit.

From this previous work \cite{sabelhaus_-situ_2022} \cite{jing2022safe}, we have evidence showing that a linear model to describe the SMA temperature dynamics is sufficiently accurate for our application. 
At time step $k$, we use $T_{k+1} = (\alpha_1(T_k - T_{amb}) + \alpha_2u_k)d_t + T_k$ as the discrete-time SMA dynamics, where $T_k$ stands for the temperature of the SMA, $u_k$ for the control input (PWM value) applied to the SMA, and $d_t$ is the time step for the control cycle. 
The three linear system characteristic coefficients are $T_{amb}$ for the ambient temperature constant, $\alpha_1$, and $\alpha_2$. 
For the simplicity of the following equations, we can rewrite the dynamics as $T_{k+1} = a_1T_k + a_2u_k + a_3$, where $a_1 = \alpha_1d_t+1$, $a_2=\alpha_2/d_t$ and $a_3=-\alpha_1T_{amb}d_t$. 
To calibrate this model, we collected SMA temperatures given random PWM input and created a 10-minute dataset for linear data fit to identify the parameters of the SMA. 
The corresponding coefficients were $T_{amb}=35.16, \alpha_1 =-0.079, \alpha_2 =29.22$, with the control period $d_t = 0.2$ secs.

In the one-step predictive supervisory controller, we limit the maximal control input so that the predicted next step temperature never exceeds the desired temperature ($T^{SET}$). 
The maximal control input is $u^*_k = \frac{1}{a_2}(T^{SET}-a_1T_k-a_3)$. 
To compensate for the modeling error, we apply a discount factor $\gamma \in (0,1]$ to $u^*_k$. 
However, the discount factor shifts the equilibrium point of the closed-loop system. 
So, we further adjust the equilibrium to $T^{SET'} = (1/\gamma - a1((1-\gamma)/\gamma))T^{SET} - a_3((1-\gamma)/\gamma)$ to make sure the actual temperature settles at $T^{SET}$. 
In this case, the updated maximal control input is $u^*_k = \frac{1}{a_2}(T^{SET'}-a_1T_k-a_3)$. 
The actual control input is $u_k = min(\hat{u}_k, \gamma u^*_k)$, where $\hat{u}_k$ is some nominal PWM input signal.
For our experiments, we chose $\gamma=0.15$.




\begin{algorithm}[h]
\For{$v\ (cycle) \leftarrow 1$ \KwTo $V$}{
    Initialize $heatedFlag,coolingFlag \leftarrow false$\;
    Initialize $heatedTime$ to zero\;
    \For {each step $k$}{
        $T_k \leftarrow readSensorData(), \hat{u}_k \leftarrow 0.5$ \;
        \If{$abs(T_k-T^{SET})<8$}{
            \If{ heatTimeFlag $is$ $false$}{
             $heatedFlag\leftarrow true$\;
             start timing $heatedTime$\;
             }
             \Else{
                \If{$heatedTime \geq 20 sec.$}{
                $coolingFlag\leftarrow true$\;
                $heatedFlag\leftarrow false$\;
                }
             }
        }
        \If{$coolingFlag$}{
            $\hat{u}_k \leftarrow 0$\;
            \If{$T_k<T^{cool}$}{
            $v=v+1$ //start next actuation cycle \; 
            }
        }
        $u^*_k \leftarrow predictMaximumInput(T_k)$ \; 
        $u_k = min(\hat{u}_k, u^*_k)$ \; 
    }
}
\caption{\bf{$\mathbf{C_1}$, Temperature-based Input Generator}} \label{alg:Algorithm-temperature}
\end{algorithm}

\begin{algorithm}[h]
\For{$v\ (cycle) \leftarrow 1$ \KwTo $V$}{
    Initialize $heatingFlag,coolingFlag \leftarrow true,false$\;
    Initialize $heatingTime, coolingTime$ to zero\;
    start timing $heatingTime$\;
    \For {each step $k$}{
        $T_k \leftarrow readSensorData(),\ \hat{u}_k\leftarrow 0.5$ \;
        \If{$heatingTime \geq 45 sec.$}{
        $coolingFlag\leftarrow true$\;
        start timing $coolingTime$\;
        }
        \If{$coolingFlag$}{
            $\hat{u}_k \leftarrow 0$\;
            \If{$coolingTime \geq 65 sec.$}{
            $v=v+1$ //start next actuation cycle \; 
            }
        }
        $u^*_k \leftarrow predictMaximumInput(T_k)$ \; 
       $u_k = min(\hat{u}_k, u^*_k)$ \; 
    }
}
\caption{\bf{$\mathbf{C_2}$, Time-based Input Generator}} 
\label{alg:Algorithm-time}
\end{algorithm}


\subsection{Input Profile Generator Controllers for Testing}

Our tests seek to mimic representative actuation profiles for SMAs in soft robots, in order to fatigue the actuators under different heating-cooling trials.
To do so, we chose two different options for the nominal control signal $\hat{u}_k$ in our control architecture (Fig. \ref{fig:system_ach}), denoted ``input profile generators'' $C_1$ and $C_2$.

Operation of soft robots that utilize artificial muscles can occur with small or large intervals between actuation periods. 
In the first use case, we can assume that between cycles of actuation, the SMAs will have enough time to cool down to a set temperature: this is $C_1$, the setpoint-based input sequence. 
In the second use case, we could instead assume that the SMAs are not given the necessary time to cool down to a set temperature but instead have a finite time to cool down: this is $C_2$, the fixed-time-based input sequence. 
During practical use, the robot will experience different heating and cooling intervals over time, and so our fatigue life limit can take the more conservative of the two results.

The $C_1$ input generator creates the sequence $\hat{u}_k$ via a set of temperature setpoints, calculated via Algorithm \ref{alg:Algorithm-temperature}. 
In this approach, the SMA is heated at a 50\% duty cycle $\hat{u}_k=0.5$ until the desired limit temperature was reached within some tolerance ($T_k - T^{SET} < 8^\circ C$). 
The temperature is then maintained by the supervisor for 20 seconds, after which, heating stops ($\hat{u}_k$ = 0.0) until the cooling temperature of $35^\circ$C is reached.
This cycle would then be repeated for as many trials as were desired.

The $C_2$ input generator creates the sequence $\hat{u}_k$ via a set of fixed-time transitions, calculated via Algorithm \ref{alg:Algorithm-time}.
In this approach, the SMA is heated at 50\% duty cycle for 45 seconds, subject to the supervisor's action.
After the 45 seconds had passed, the SMA would be allowed to cool for 65 seconds, marking the end of a cycle. 

\begin{figure}[t]
    \centering
    \includegraphics[width=1\columnwidth]{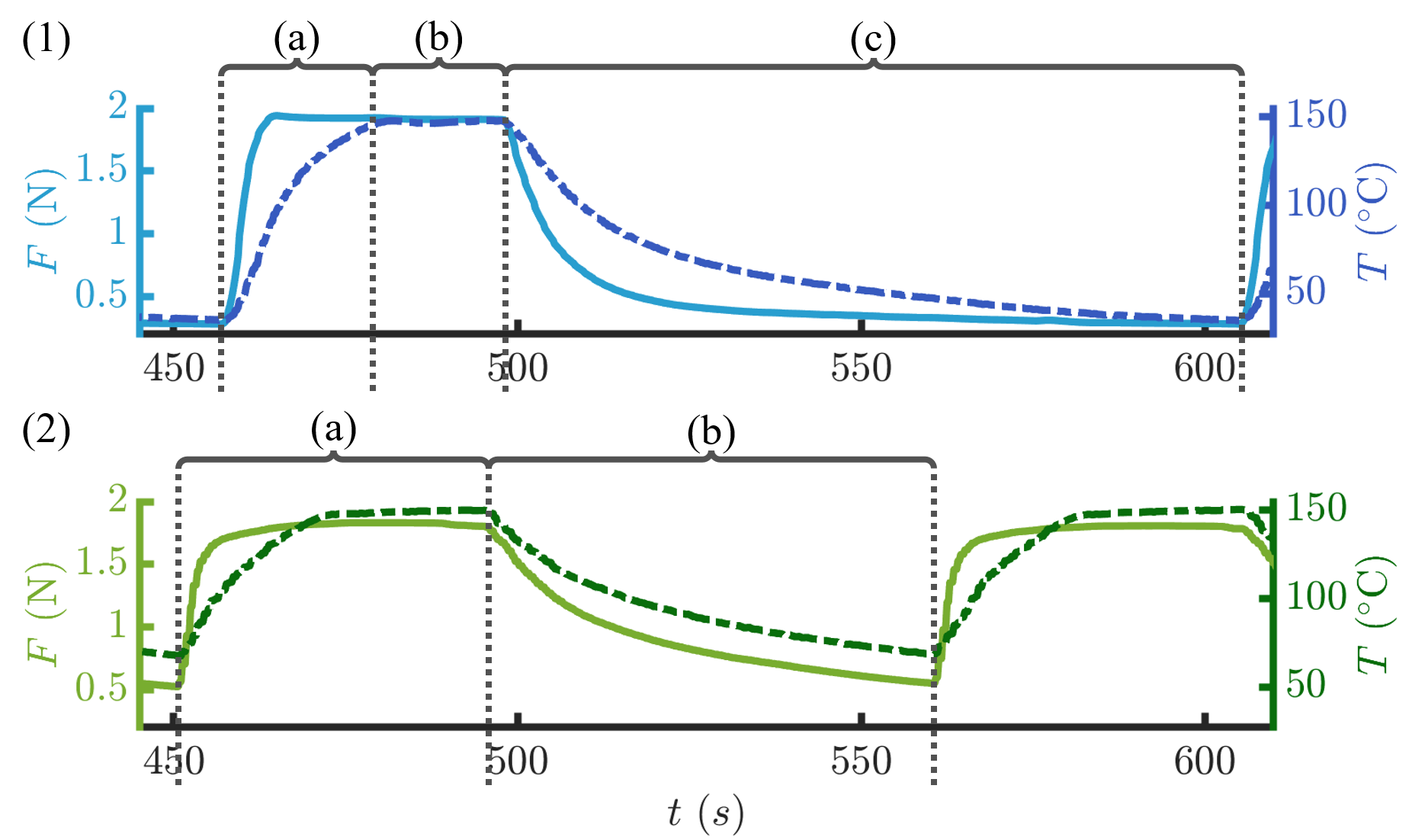}
    \caption{Example results from controller $C_1$ and controller $C_2$ with temperature maximum of $150^\circ$C. Plot (1) shows reaching the limit temperature (1a), holding the limit temperature for 20 seconds (1b), and then cooling to $35^\circ$C (1c) as set by $C_1$. Plot (2) shows the fixed 45 seconds of heating (2a) and 65 seconds of cooling (2b) as set by $C_2$.}
    \label{fig:sample_curves}
    \vspace{-0.3cm}
\end{figure}

The key differences between the approaches are related to the temperature range the SMAs would experience within each cycle and the total time the SMA would experience its set limit temperature.
Qualitatively, $C_2$ sequence cycled the actuators faster, with less cool-down between cycles, than $C_1$.
Representative cycles from each are shown in Fig. \ref{fig:sample_curves}.



\subsection{Experimental Procedures}

Our experimental procedure applies each framework to collect a large dataset of forces and temperatures.
For each trial, we select a maximum temperature $T^{SET}$, fabricate a fresh SMA bundle, and assemble the limb into the force-plate test setup.
We then execute the control framework with either $C_1$ or $C_2$ for a total of 100 cycles, measuring the blocked bending force at the limb's tip.
This process is repeated through a set of temperatures for each controller.

\begin{figure*}[t]
    \centering
    \includegraphics[width = \textwidth]{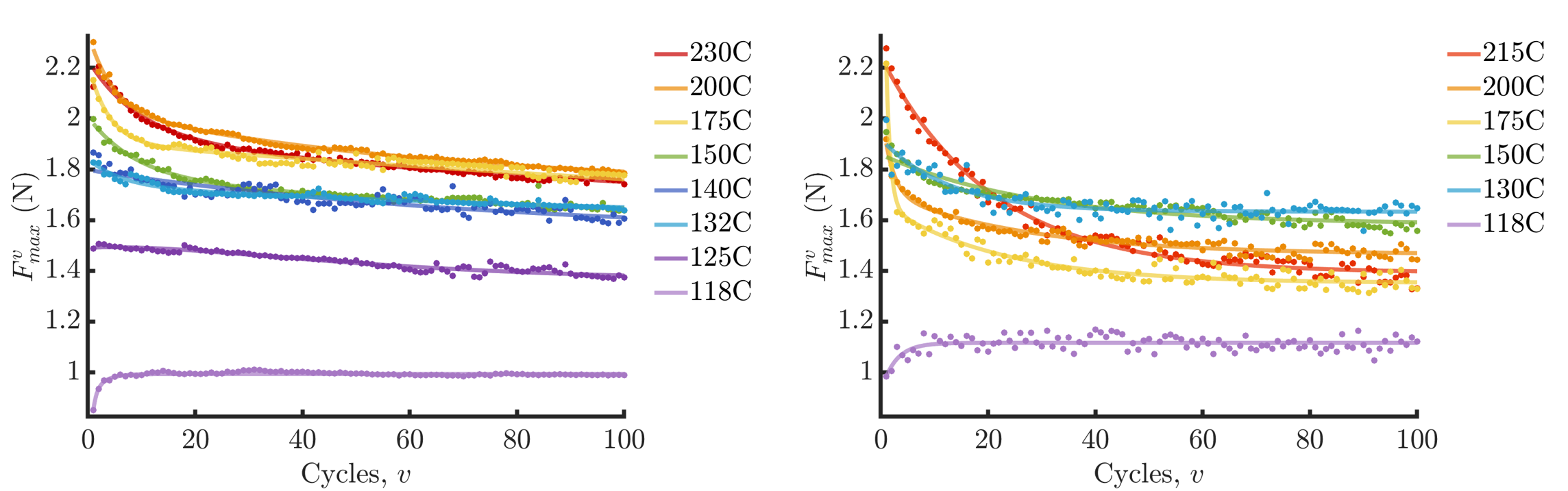}
    \caption{Maximum force output of SMA per cycle with trials performed at different limit temperatures using two controller approaches. The plot on the left shows the results when utilizing a temperature setpoint input generator $C_1$ for 100 cycles. The plot on the right shows the results when utilizing a fixed time input generator $C_2$ through 100 cycles.}
    \label{fig:plotAllRuns}
    \vspace{-0.2cm}
\end{figure*}

Each trial generates a set of datapoints $\{F_t, T_t\} \in \mathbb{R}^{2\times N}$ for $N$ timesteps, e.g., $\{F_0, T_0, F_1, T_1, \hdots, F_N, T_N \}$.
Within this trial, there are $v = 1 \hdots V$ heating cycles, where for all our tests, $V=100$.
Our supervisory control system prevents the SMA from exceeding a set maximum temperature during this process.
For both methods, we calculate a maximum force $F_{max}^v$ for trial $v$ by taking the max force measurement over a window of values corresponding with cycle $v$ in the dataset.
For the fixed-time test, we select the $v$-th window in the force data starting at the start time $t=k$, identified using the controller's output trajectory, until the end of the heating period at $t=k+K$, where $K=45$ seconds.
Within this set of $\{F_k, \hdots, F_{k+K}\}$, we manually filter any outliers in the data (see Discussion), then take $F_{max}^v = \text{max}(\{F_k, \hdots, F_{k+K}\})$.

Similarly, for the temperature setpoint test, we selected the window indices $k$ and $K$ based on a trace of our controller outputs, where $k$ was such that $T_k > T^{set} - 8^\circ $C.
The end of the window, $K$, was selected at the next timepoint where the temperature decreased back down, i.e., $T_{k+K} < T^{set} - 8^\circ $C.
We filtered outliers by removing the first portion window where we observed transients, e.g., $k' = k + (K/2)$ to remove half the window.
This procedure was performed by examining the controller's oscillations at each setpoint.
Then, $F_{max}^v = \text{max}(\{F_{k'}, \hdots, F_{k' + K}\})$.

%% file: P3_Results.tex
\section{Results}
With the goal of maximizing a consistent force output, we compared these two methods of fatigue testing, measuring the complex relationship between temperature, maximum force over time, and cyclic loading sequence. 
The raw data from fourteen of these tests is shown in Fig. \ref{fig:plotAllRuns}.
Each $T^{SET}$ was chosen empirically at regularly-spaced temperature intervals, starting at the values from our prior work \cite{sabelhaus_safe_2022,jing2022safe} and increasing or decreasing until we observed qualitative changes in the curves.


\subsection{Long-Life Force and Temperature Prediction}

To draw generalizable conclusions, we fit each trajectory to an exponential decay.
This curve fit serves two purposes: first, we may better visualize the overall trend for each temperature trial, but second and more important, we can predict the long-term maximum force output, $F_{max}^\infty$, from the curve fit.
The exponential-decay fits were chosen empirically for their low root-mean-square error (RMSE) versus our datasets; future work will examine if a more physically-motivated model is appropriate. 

To do this fit for each of the temperatures, where $j=1,\hdots, J$ and $J=8$ for the temperature setpoint method and $J=6$ for the fixed-time method, we fit two variations on an exponential decay and chose the version with the lower RMSE:
\begin{align}
F_{max}(v)^j & = a_j e^{-b_j v}+c_j, \\
F_{max}(v)^j & =a_j e^{-b_j v}+ d_j e^{-g_j v}+c_j
\end{align}

\noindent These fits are plotted alongside the discrete $F_{max}^v$ points in Fig. \ref{fig:plotAllRuns}. 
We observe that both the $C_1$ and $C_2$ trends generally showed continually decreasing force, implying that more cycles would show more fatigue or degradation.

\begin{figure}[t]
    \centering
    \includegraphics[width = \columnwidth]{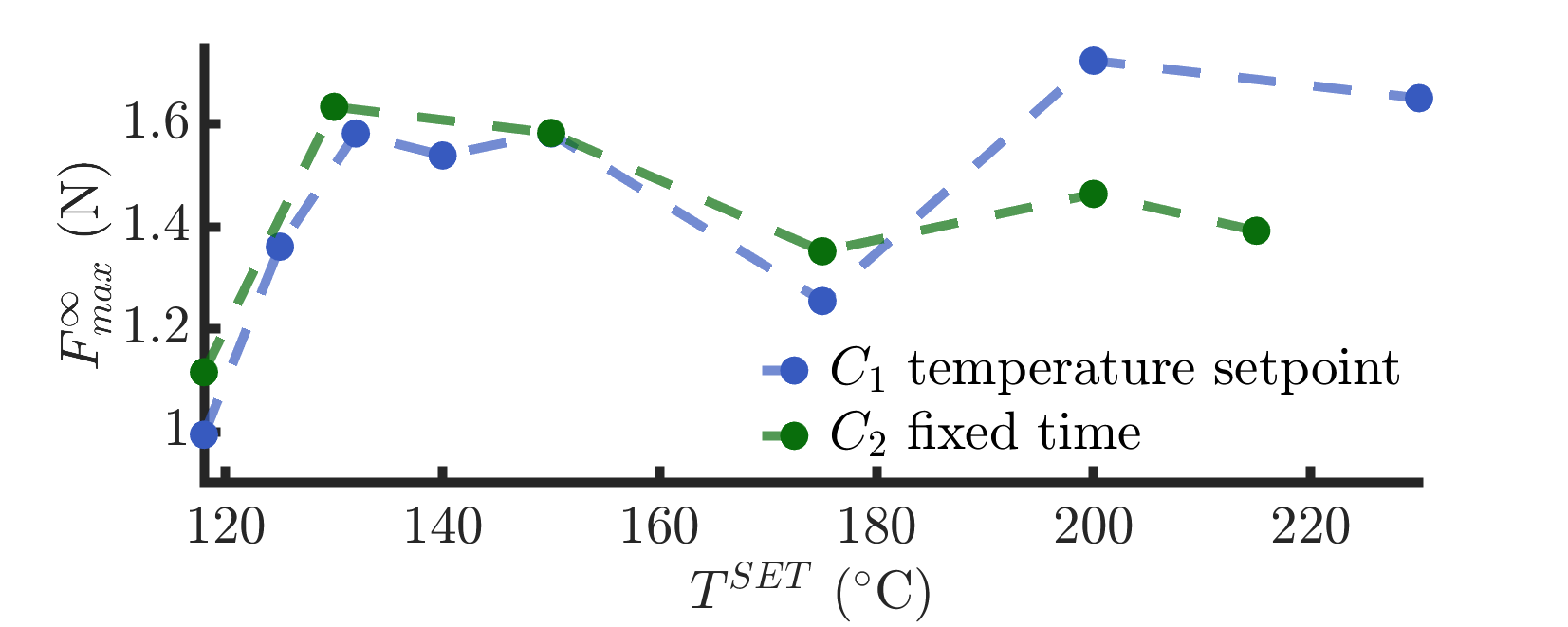}
    \caption{Predicted long term force output at different temperatures for both controller methods. Based on the data previously gathered, this plot contains the predicted long term equilibrium force output for both controller types: temperature setpoint ($C_1$) and fixed time ($C_2$).}
    \label{fig:F_infinity}
    \vspace{-0.3cm}
\end{figure}

From the equations of best fit, we propose that $F_{max}^\infty = c_j$.
This constant offset would be the predicted settling value as the number of cycles $V \rightarrow \infty$.
We plotted each of the fourteen $F_{max}^\infty$ points versus the temperature limit that generated them (Fig. \ref{fig:F_infinity}).

Our data suggest a concrete answer to the core question of SMA lifetime: what ranges of maximum temperatures ensure repeatable behavior?
We first observe that similar long-life forces occur across both methods for temperatures in the range of 118$^{\circ}$C to 175$^{\circ}$C, at a maximum around 1.5N - 1.6N.
Above that temperature, results are inconsistent between the two actuation profiles, yet the $F_{max}^\infty$ forces do not significantly exceed those of the lower temperatures in either profile.
We therefore propose a conservative approach: the maximum-life temperature should be the lowest-temperature inflection point between the two profiles.
For our SMAs, this maximum-life temperature is $T^{SET}=140^\circ$ C.

\subsection{Validation of Long-Life Maximum Forces}

\begin{figure}[t]
    \centering
    \includegraphics[width = \columnwidth]{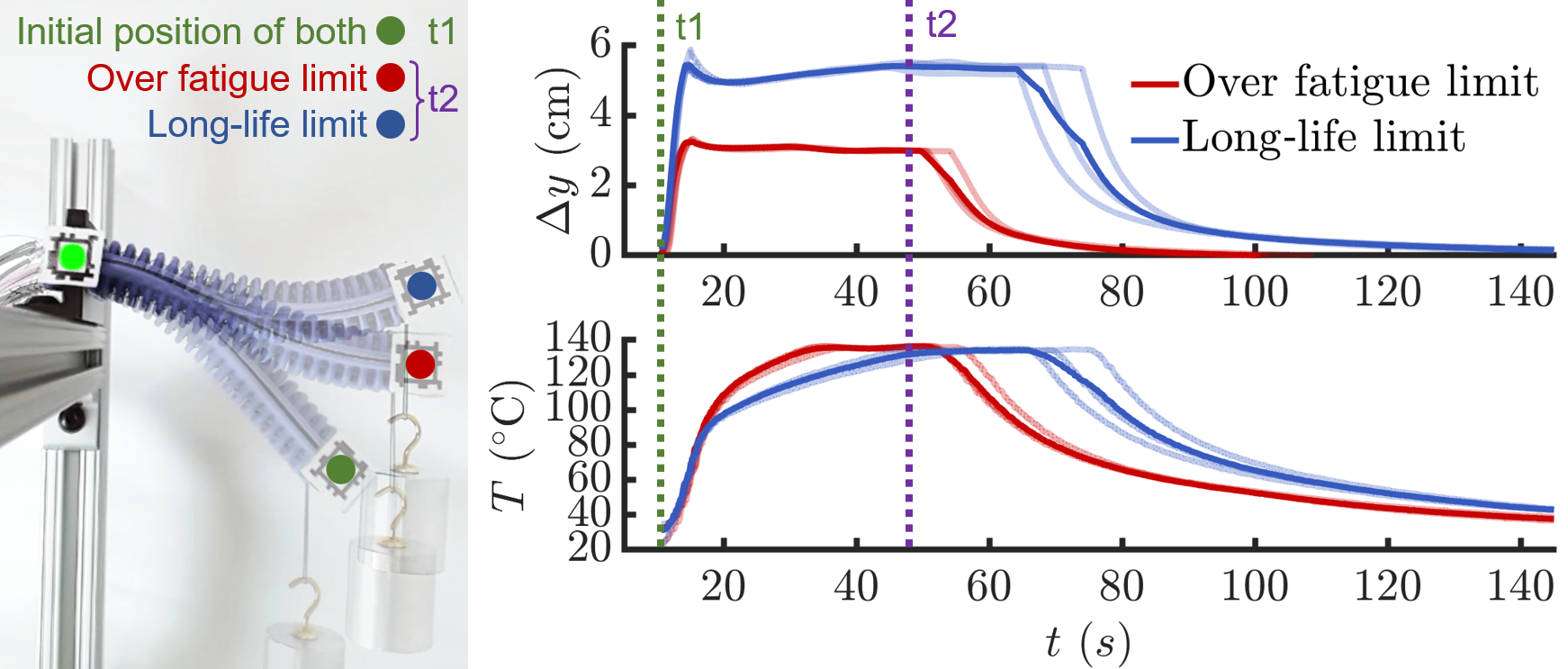}
    \caption{A displacement test with a 50g load demonstrates that an SMA controlled at its long-life limit consistently generates larger motions. A prototype fatigued above the limit displaces to the red dot position (left), whereas the long-life limb raises the weight to the blue dot. Each of two wires were actuated per the $C_1$ profile (right), with snapshots taken at a timepoint of the same wire temperature (t2) for comparison. Three cycles are shown per test; solid lines are average values.}
    \label{fig:displacement}
    \vspace{-0.3cm}
\end{figure}

To validate our approach, we placed two prototypes from this experiment in a free motion test of lifting a weight (Fig. \ref{fig:displacement}, left).
We compared the SMA that had been cycled with $T^{SET}=140^\circ$ C versus the SMA with $T^{SET} = 230^\circ$ C, both from the $C_1$ controller (temperature setpoint), which had the noisy long-life predictions at higher temperatures.
Each limb with corresponding SMA bundle was turned on its side, with a 50g mass attached to the tip, then subjected to $C_1$ again at $T^{SET}=140^\circ$ C.
We cycled each SMA an additional 50 cycles in this arrangement.

The limb with SMAs fatigued at 230$^{\circ}$C experiences a significantly lower displacement when compared to a limb fatigued at 140$^{\circ}$C, seen in three representative cycles' computer vision tracking data in Fig. \ref{fig:displacement}. 
This displacement corresponds to SMA force, which degraded by a factor of two for the SMA muscle that had been kept above our maximum long-life temperature.
These results verify that our conservative $T^{SET}$ maintained the highest-observed long-life force output limit.


%% file: P4_Discussion.tex
\section{Discussion}

The results in this article show that a soft robot, powered by SMA actuators, can be controlled to consistently apply high forces and displacements at long cycle lifetimes (150+ cycles).
This work is one of the first results, in terms of design and control, that induced consistent output forces from an SMA artificial muscle in-situ. 
The experiments emphasize that the stimulus-response behavior of SMA artificial muscles does not exist in isolation, but is influenced by their robot's design, sensing, actuation, and control. 
Our method specifically lets us correlate in-situ sensor readings with muscle lifetime within the robot itself, a key to moving these robots into real-world applications.

\subsection{Implications of Cycle Profiles versus Lifetime}

In Fig. \ref{fig:F_infinity}, there is a clear inflection point where as temperature increases, force decreases. 
Beyond this point, the use case begins to have a greater influence on the behavior, showing inconsistent trends from the $C_1$ input sequence generator (longer holds with more cooling) versus the $C_2$ inputs (faster cycling, less cooling). 
Consequently, our recommendation is to be conservative when the robot's application is not yet known and to choose the temperature limit at the lower-temperature inflection point between the two approaches.

Additionally, there are diminishing returns beyond the temperature inflection point, where even in an ideal scenario where the SMA can cool down to 35$^{\circ}$C, achieving an addition 0.1N more force requires more than 50 degrees of heating. 
This is expected given the materials science behind the shape-memory effect, and our method provides a quantifiable approach for designers to judge how to set these limits specifically.

\subsection{Limitations of Methodology}

Despite the clear and useful conclusions that arise from our methodology, there are limitations to the approach.
First, inconsistencies in both fabrication by-hand and variation in the off-the-shelf SMAs themselves will inevitably limit the ability to draw larger conclusions about cycle life for forces above $F_{max}^\infty$.
In particular, though some of the observed forces were larger at temperatures above the fatigue limit, our data was not sufficient to reliably predict forces at these higher temperatures.
Fortunately, this work is not directed towards such a nuanced relationship, and instead proposes a reliable lower bound on maximum consistent forces.

Second, our two input sequences (heating cycle profiles) were chosen based on past experience rather than a rigorous process.
We propose that $C_1$ and $C_2$ reasonably reflect different use case scenarios the limb could encounter in real-world settings, but other loading sequences may show a different relationship or break with temperature-dependence entirely.


Finally, this approach uses a curve fit to predict a long-lifetime maximum force, which inherently introduces assumptions that may not hold.
The two different exponential decays were chosen as fitting curves, empirically, based on quality of fit (via RMSE), rather than first-principles reasoning.
These curve fits were strikingly consistent in their predictions at low-to-moderate temperatures in Fig. \ref{fig:F_infinity}, and those predictions held in a hardware experiment.
So although our approach is useful for estimating the inflection point in terms of temperature, different learned models would likely generate different fatigued force predictions at higher temperatures.
Our general approach of identifying a conservative limit via an inflection point may still remain viable even in those situations.




%% file: P5_Conclusion.tex
\section{Conclusion}
This article proposes a framework for design, sensing, and control that maintains long-life force and displacement for shape memory alloy (SMA) actuators in soft robots.
Our result demonstrates that temperature sensing with feedback may be sufficient to prevent functional fatigue in these artificial muscles in practical soft robotics tasks.
The framework was successfully demonstrated on a coiled Nitinol SMA actuator with a nominal transition temperature of $90^\circ$C, where a temperature limit in the range of 130$^{\circ}$C to 150$^{\circ}$C was ideal (via our experimental methodology) for the muscle's cycle life while maximizing consistent force output.


This article is one of the first practical attempts to quantify SMA lifespan versus control, but many open questions remain. 
Future work will focus on evaluating lifetimes and consistent forces including strain, as this article reports blocked force only, as well as internal material phase, as this article only senses temperature.

More importantly, future work will seek to implement this method on large, high-force, high-displacement soft robot limbs, with sensing and estimation of contact.
Maximizing these forces while maintaining a consistent relationship between temperature, control input, and output force will be key to estimating a soft SMA-powered robot's pose and applied forces.
Using these results, soft robots may one day be able to interact meaningfully with their environments at human-size scales.